\documentclass[lettersize,journal]{IEEEtran}
\pdfoutput=1
\usepackage{amsmath,amsfonts}
\usepackage{algorithm}
\usepackage[noend]{algpseudocode}

\usepackage{array}
\usepackage[caption=false,font=normalsize,labelfont=sf,textfont=sf]{subfig}
\usepackage{textcomp}
\usepackage{hyperref}
\usepackage{stfloats}
\usepackage{booktabs}
\usepackage{url}
\usepackage{verbatim}
\usepackage{graphicx}
\usepackage{caption}
\usepackage{threeparttable}
\usepackage{cite}
\usepackage{xcolor}
\def\BioMime/{BioMime}

\begin{document}

\title{Conditional Generative Models for Simulation of EMG During Naturalistic Movements}
\author{Shihan Ma\textsuperscript{*}, Alexander Kenneth Clarke\textsuperscript{*},  Kostiantyn Maksymenko, Samuel Deslauriers-Gauthier, Xinjun Sheng,~\IEEEmembership{Fellow,~IEEE}, Xiangyang Zhu,~\IEEEmembership{Fellow,~IEEE}, Dario~Farina,~\IEEEmembership{Fellow,~IEEE}}

\maketitle

\begin{abstract}
Numerical models of electromyographic (EMG) signals have provided a huge contribution to our fundamental understanding of human neurophysiology and remain a central pillar of motor neuroscience and the development of human-machine interfaces. However, whilst modern biophysical simulations based on finite element methods are highly accurate, they are extremely computationally expensive and thus are generally limited to modelling static systems such as isometrically contracting limbs. As a solution to this problem, we propose a transfer learning approach, in which a conditional generative model is trained to mimic the output of an advanced numerical model. To this end, we present BioMime, a conditional generative neural network trained adversarially to generate motor unit activation potential waveforms under a wide variety of volume conductor parameters. We demonstrate the ability of such a model to predictively interpolate between a much smaller number of numerical model's outputs with a high accuracy. Consequently, the computational load is dramatically reduced, which allows the rapid simulation of EMG signals during truly dynamic and naturalistic movements.
\end{abstract}


\thanks{D. Farina, S. Ma and A. K. Clarke are with the Department of Bioengineering, Imperial College London, UK. S. Ma is also with the State Key Laboratory of Mechanical System and Vibration, Shanghai Jiao Tong University, Shanghai, China, along with X. Sheng and X. Zhu. X. Sheng and X. Zhu are also with the Meta Robotics Institute, Shanghai Jiao Tong University, Shanghai, China. K. Maksymenko and S. Deslauriers-Gauthier are with the company Neurodec, based in Sophia Antipolis, France. S. Deslauriers-Gauthier is also with the Inria Centre at Université Côte d'Azur, Nice, France.}

\thanks{This study is supported by the National Natural Science Foundation of China (Grant No. 91948302, 52175021), the European Research Council Synergy Grant NaturalBionicS (contract 810346), the EPSRC Transformative Healthcare, NISNEM Technology (EP/T020970), and the BBSRC, ``Neural Commands for Fast Movements in the Primate Motor System'' (NU-003743). D. Farina (d.farina@imperial.ac.uk) and X. Zhu (mexyzhu@sjtu.edu.cn) are the corresponding authors. \textsuperscript{*}The first two authors contributed equally to this work.}

\section{Introduction}
\IEEEPARstart{B}{iophysical} simulations are a cornerstone of modern biomedical research and engineering, allowing initial explorations of experimental hypotheses and fast iterations of designs prior to physical implementation \cite{gerstner2012theory, halilaj2018machine}. Decades of continuous developments have seen such models go from a few equations, such as in Hodgkin and Huxley's hugely impactful work on spiking neurons \cite{hodgkin1952quantitative}, to highly complex physics engines and large numerical models with thousands of individual parameters \cite{breakspear2017dynamic}. The rapid expansion in the complexity and fidelity of biophysical simulations has played a major role in advancing their corresponding domains \cite{baby2021convolutional}, and has even generated entirely new avenues of investigation, for example, neurophysiological source reconstruction and embodied artificial intelligence \cite{fuchs1998improved, caggiano2022myosuite}. The importance of accurate biophysical simulations is particularly important for research which involves electromyography (EMG) time series processing, a signal finding increasing use in medical and consumer applications due to its ease of recording and rich descriptions of muscle activities\cite{wang2021novel}. The simulated data forms a ground truth for validating decomposition algorithms and solving inverse modelling problems\cite{farina2017man, merletti1999modeling, clarke2023deep}. The data can also be used to augment experimental data for improving motor control \cite{maksymenko2023myoelectric}.

Despite the successes of such biophysical models, the improved accuracy has come with a corresponding increase in the associated computational burden\cite{baby2021convolutional}. Accurate simulations of EMG usually involve numerical modelling of neural source propagation through an anatomically-precise volume conductor, which is described by Poisson's equations and solved using a finite element method (FEM)\cite{maksymenko2023myoelectric}. This approach excels when the volume conductor is fixed. However, when the underlying parameters of the biophysical system are evolving, for example in a moving forearm, the structural mesh and forward model must be repeatedly recalculated as the 3D model changes\cite{ma2020subject}. The dynamic movement must be divided into a sequence of stationary stages \cite{pereira2019anatomically}, with the temporal resolution heavily limited by computational demands. A lack of EMG simulations during dynamic and naturalistic movements remains a major bottleneck in the development of EMG signal processing methods, which move beyond the current paradigm of isometric contractions\cite{holobar2021noninvasive}.

One computationally-feasible alternative approach to modelling a dynamically evolving volume conductor is to build generative models which can be trained to mimic a numerical model's outputs across a variety of system conditions. If a model could accurately interpolate between the outputs of numerical simulations with very different system conditions, then the required number of FEM runs would be drastically reduced. In this study, we demonstrate that such interpolation accuracy is possible by proposing BioMime, a hybrid conditional generative neural network with an encoder-decoder structure. BioMime is trained adversarially within a robust transfer learning pipeline\cite{wang2022reinforcement, yan2016attribute2image, gebauer2022inverse}. We show that BioMime can accurately fill in the gaps between the system conditions of an advanced FEM model for generating surface EMG signals sensed by electrodes on the human forearm. As a demonstration of the potential applications of such efficiency, we provide an example of using BioMime to simulate surface EMG signals during human hand, wrist, and forearm voluntary movements.

\begin{figure*}[h]
  \centering
  \includegraphics[width=1.6\columnwidth]{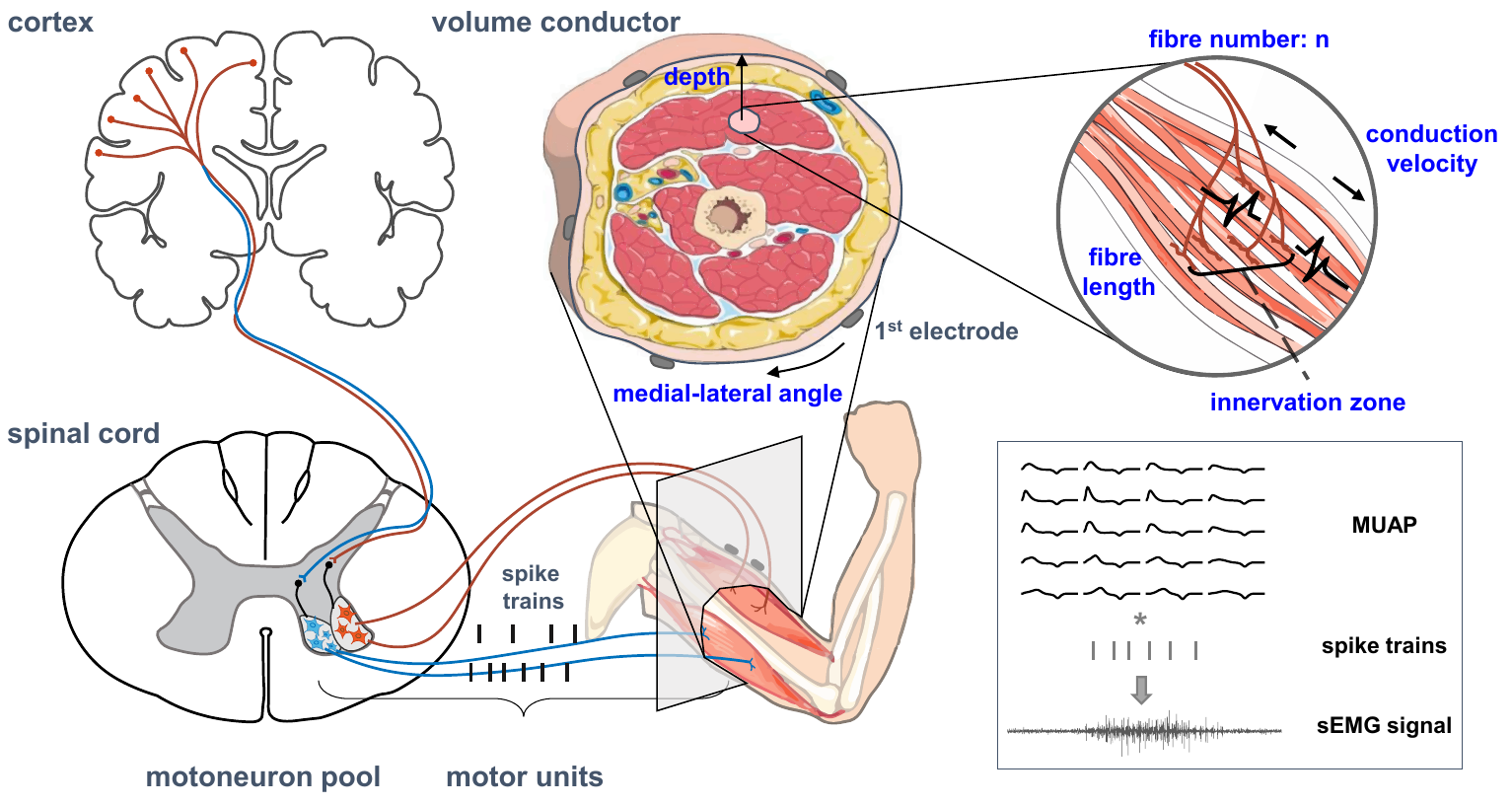}
  \caption{Capturing a complex surface EMG simulation.
  Surface EMG signals represent the myoelectric output of motor neuron activities from the spinal cord and are detected by an array of electrodes placed on the skin above the muscles. Each spike from a specific motor neuron activates an associated pool of muscle fibres, where the combination of the motor neuron and the muscle fibres it controls is called a motor unit (MU). The electrical potential field caused by the muscle activities is filtered by the volume conductor and sensed at the surface electrodes as the motor unit action potential (MUAP). Simulated MUAPs are convolved with the desired spike activities and then summed (with optional added Gaussian noise) to build the simulated surface EMG. The bulk of MUAP variance is explained by six specified generative factors (printed in blue). These six factors are the specified system conditions in the case of MUAP simulation.
  }
  \label{fig:grand}
\end{figure*}

\section{Preliminaries}

\subsection{The Generative Model of Surface EMG}
Modern methods of surface EMG simulation track closely with the neuroscience and biophysics in the peripheral motor and musculoskeletal systems (Fig. \ref{fig:grand}). A single  motor neuron and the muscle fibres it activates constitute a motor unit (MU), the smallest functional unit of the motor system\cite{merletti2016surface}. When the muscle fibres receive the neural inputs from the motor neurons in the form of spike trains, the generated current source action potentials begin at the muscle fibre end plates, collectively called the innervation zone (IZ), before propagating longitudinally along the fibre in both directions\cite{farina2004surface}. The waveform created by the summed action potentials from all the fibres in a single MU is named the motor unit action potential (MUAP), which arrives at the sensing electrodes after propagating through the intervening tissues, called the volume conductor. Surface EMG signal represents the summed MUAPs from all active MU sources with additional enviromental noise.

Assuming conditions of isometric contractions, this can be represented mathematically as a convolutional mixture\cite{negro2016multi}, where an $N$-sized vector of spiking sources $\boldsymbol{s}$ is being sensed by a $M$-sized vector of sensing electrodes $\boldsymbol{r}$ at time $t$ after attenuated by a finite impulse response filter $\boldsymbol{H}(l)$ of time support $L$ and dimension $M \times N$ per time instant $l$:

\begin{equation}
    \boldsymbol{r}(t) = \sum_{l=0}^{L-1}\boldsymbol{H}(l)\boldsymbol{s}(t-l)+\boldsymbol{z}(t)
    \label{eq:1}
\end{equation}
where $\boldsymbol{z}(t)$ is an $M$-sized vector of additive noise at time $t$.

If the contraction is non-isometric, $\boldsymbol{H}(l)$ will change continuously throughout $L$ with the changing geometries of the volume conductor. If the rate of change of the volume conductor is much longer than $L$, as is usually the case, then an approximation can be made wherein the volume conductor conditions at the start of the MUAP waveform are assumed to persist throughout it. This simplifies the simulation of the MUAP finite impulse response filter $\boldsymbol{H}$ to a three dimensional MUAP template $\boldsymbol{x}$ of time support $T$ and spatial support $H \times W$, which is then convolved in the time dimension with the spiking time series $\boldsymbol{s}$ to generate EMG signals.

\subsection{Simulation by Finite Element Method}

Equation \ref{eq:1} motivates the division of EMG simulation into two components: simulation of the motor neuron spike trains $\boldsymbol{s}$ and simulation of their associated MUAP templates $\boldsymbol{x}$. Simulation of the spike trains is computationally simple and is based on established models of motor neuron spiking and recruitment behaviour\cite{fuglevand1993models, caillet2022estimation}. Simulation of the MUAP templates is a far more complex operation as it involves modelling of the bioelectric potential field propagation as they are attenuated by the materials of the volume conductor\cite{maksymenko2023myoelectric}. This can be represented mathematically as Poisson's equations with Neumann boundary condition\cite{farina2004advances}:

\begin{equation}
  \begin{split}
  \nabla \cdot (\boldsymbol{\sigma} \nabla \phi) = -I ~~\text{   in } \boldsymbol{\Omega} \\
  \dfrac{\partial \phi}{\partial \boldsymbol{n}} = 0 ~~\text{ on } \partial \boldsymbol{\Omega} ~
  \end{split}
  \label{eq:2}
\end{equation}
where $\boldsymbol{\Omega}$ is the domain of definition of the three-dimensional volume conductor with boundary $\partial \boldsymbol{\Omega}$, $\phi$ is the potential, $I$ is the current density source, $\boldsymbol{\sigma}$ is a conductivity tensor and $n$ is the versor normal to $\partial \boldsymbol{\Omega}$. The boundary condition models the assumption that there is no current flow between the skin and air.

Whilst analytical models are available for solving equation \ref{eq:2} for simple systems, such as a cylindrical or planar volume conductors\cite{farina1999compensation, farina2004surface}, generally realistic systems require numerical modelling techniques such as FEM \cite{farina2004advances}. In FEM, the volume conductor is broken down into tetrahedral elements, which discretises equation (\ref{eq:2}) into $j$ linear equations, where $j$ is the number of mesh vertices:

\begin{equation}
\boldsymbol{A}\boldsymbol{v} = \boldsymbol{b}
\label{eq:3}
\end{equation}
where $\boldsymbol{A}$ is a symmetric and sparse $j \times j$ matrix, $\boldsymbol{b}$ is a $j$-sized vector of source information, and $\boldsymbol{v}$ is a $j$-sized vector of potential values constrained to have zero sum.

Whilst FEM modelling can accurately replicate the biophysics of an anatomically-complex volume conductor\cite{lowery2004volume, pereira2019anatomically, maksymenko2023myoelectric}, the solution is only correct for the current mesh. This means that any mesh deformation, such as during a non-isometric contraction, requires the recalculation of equation \ref{eq:3}. This represents a massive computational load for systems with a high number of vertices. This motivates the need for a generative model which can predictively interpolate between a few FEM solutions. Such a generative model could reduce the frequency of FEM recalculations when simulating dynamic systems and make the simulations computationally feasible.

\subsection{Transfer Learning with Conditional Generative Models}
Generative neural networks have demonstrated a remarkable ability to synthesise a wide range of realistic data\cite{wen2023rapid, song2019generative}.
The objective of this study is to transfer the knowledge from the teacher numerical model to the pupil conditional generative network\cite{wang2021intelligent}. Specifically, we wish to train a conditional generative model on MUAP templates $\boldsymbol{x}$ that are generated by FEM models with a simplified version of system conditions $\boldsymbol{c_0}$, such that the conditional generative model can take a new input of system conditions $\boldsymbol{c_s}$ to generate a new MUAP template $\boldsymbol{\tilde{x}}$.

\begin{figure}[h]
  \centering
  \includegraphics[width=0.8\columnwidth]{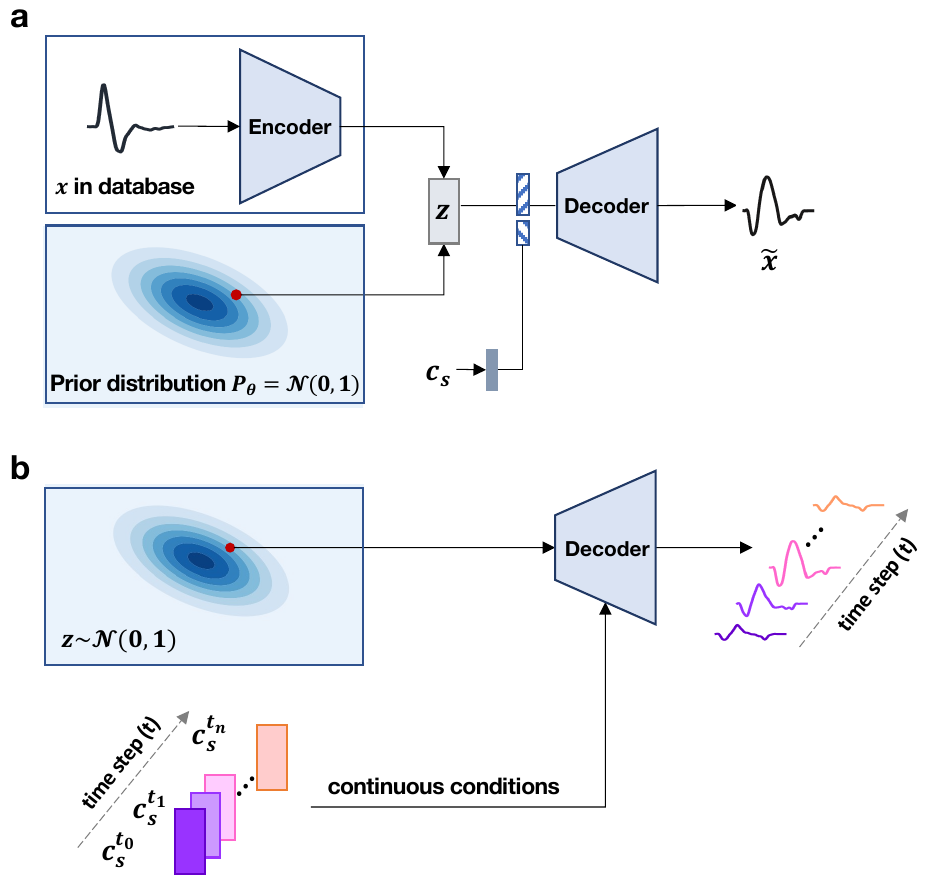}
  \caption{BioMime, a conditional generative neural network. a, Architecture of the generator in the BioMime model. The generator in BioMime consists of an encoder and a conditional decoder. The decoder takes either an encoded input MUAP or a standard normal-distributed latent as an input $\boldsymbol{z}$ and outputs a new sample $\boldsymbol{\tilde{x}}$. The output $\boldsymbol{\tilde{x}}$ incorporates the specified system conditions $\boldsymbol{c_s}$ from the generative process. b, Rapid generation of a dynamically evolving MUAP \textit{ab initio}. A sample is taken from the prior distribution and then continuously transformed over time using a sweep of the specified system conditions. Such simulations interpolate between the FEM model's outputs and thus greatly reduce the computational burdens when the system conditions evolve.
  }
  \label{fig:biomime_concept}
\end{figure}

The model should have two modes of operation, a ``morphing'' mode in which an existing MUAP template is encoded and then adjusted with a new set of $\boldsymbol{c_s}$ conditioning variables, and an \textit{ab initio} generation mode without the requirement of any existing MUAPs. $\boldsymbol{c_s}$ includes parameters such as the source position coordinates, IZ, conduction velocity, etc. Using the set of simplified parameters $\boldsymbol{c_s}$ rather than the fully specified mesh avoids repetitive calculations as in the FEM models. As this simplification will inevitably fail to explain the full variance in $\boldsymbol{\tilde{x}}$ during the morphing mode of operation, we elected to use an encoder-decoder structure as the network architecture (Fig. \ref{fig:biomime_concept}). This allows the additional variance $\boldsymbol{c_u}$ to be explained by an encoded version of $\boldsymbol{x}$. $\boldsymbol{c_u}$ is concatenated with a new $\boldsymbol{c_s}$ before being passed to a decoder to get the desired $\boldsymbol{\tilde{x}}$. In the \textit{ab initio} generation mode, $\boldsymbol{c_u}$ is instead populated with samples from a standard normal distribution. For simplicity, we will refer to this encoder-decoder conditional generative neural network for MUAP generation as BioMime.

\begin{figure}[h]
  \centering
  \includegraphics[width=\columnwidth]{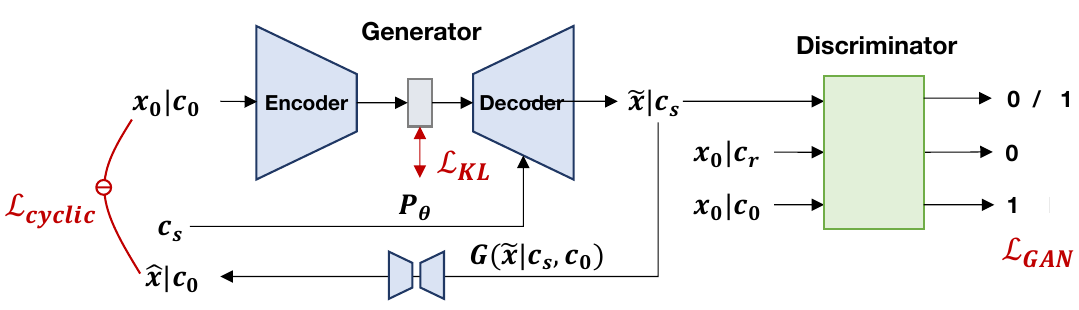}
  \caption[Architecture and loss function of the proposed model.]{Architecture and loss function of the proposed model. \BioMime/ is trained with an adversarial loss $\mathcal{L}_{GAN}$ using a discriminator, which seeks to distinguish between the generated data $\boldsymbol{\tilde{x}}$ and the outputs of the numerical simulations $\boldsymbol{x_0}$. The discriminator is also conditioned on the specified system conditions $\boldsymbol{c_s}$. This means that the generator must learn whether the sample was likely to be the result of a simulation with the specified system conditions. To allow for \textit{ab initio} generation and to stabilise training, an additional Kullback-Leibler divergence $\mathcal{L}_{KL}$ term is minimised between the encoded MUAP templates $\boldsymbol{c_u}$ and a standard normal prior. Finally, we empirically found that the addition of a cycle-consistency loss $\mathcal{L}_{cyclic}$ improved training stability and gave an increase in model performance.}
  \label{fig:biomime_loss}
\end{figure}

The encoder-decoder architecture of BioMime is trained adversarially using a conditional discriminator network-driven loss function, minimising the following loss function (Fig. \ref{fig:biomime_loss}):

\begin{equation}
  \mathcal{L}_{G}= \lambda_ 1 \mathcal{L}_{GAN} + \lambda_ 2 \mathcal{L}_{KL} + \lambda_3 \mathcal{L}_{cyclic}
  \label{eq:4}
\end{equation}

The first term $\mathcal{L}_{GAN}$ is an adversarial loss, which evaluates the network performance using the conditional discriminator. The second term $\mathcal{L}_{KL}$ is the Kullback-Leibler divergence ($\mathcal{D}_{KL}(\cdot\|\cdot)$) between the predicted distribution of the latent feature and the standard normal distribution $\mathcal{N}(\boldsymbol{0},\boldsymbol{I})$. Minimizing $\mathcal{L}_{KL}$ regularises the latent space to approach $\mathcal{N}(\boldsymbol{0},\boldsymbol{I})$. This enables the \textit{ab initio} generation mode whilst also acting as a regulariser. An additional cycle-consistency loss $\mathcal{L}_{cyclic}$ is included to improve training stability and generation accuracy, which is the mean-squared error between the input sample and a generated sample with the same system conditions.

The objective of the conditional discriminator is to differentiate the data samples produced by the generator from those in the original dataset whilst also detecting whether the samples match the specific conditions. This is done by minimising a loss formed by the addition of three binary crossentropy terms:

\begin{equation}
  \mathcal{L}_{D}= \lambda_ 4 \mathcal{L}_{CC} + \lambda_ 5 \mathcal{L}_{IC} + \lambda_6 \mathcal{L}_{Gen}
  \label{eq:4}
\end{equation}
where $\mathcal{L}_{CC}$ is the accuracy of the conditional discriminator in predicting that MUAP waveforms $\boldsymbol{x_0}$ from the FEM model with the correct set of system conditions $\boldsymbol{c_0}$ are from the training set, $\mathcal{L}_{IC}$ is the accuracy of predicting that $\boldsymbol{x_0}$ that have a randomised pairing of system paramters $\boldsymbol{c_r}$ are not from the training set, and $\mathcal{L}_{IC}$ is the accuracy of predicting that samples $\boldsymbol{\tilde{x}}$ from the conditional generative model with a paired correct system conditions $\boldsymbol{c_0}$ are not from the training set.

\section{Materials and Methods}

\subsection{Training and Validation Datasets}
To examine the ability of BioMime to accurately capture the outputs of a numerical simulation and then interpolate between the system conditions, we used an advanced FEM model of a realistic forearm anatomy, Neurodec\cite{maksymenko2023myoelectric}, to generate the dataset. For each of the 1500 MU, the MUAP templates under 256 conditions were generated, which were the combinations of four fibre densities (200, 266, 333, 400 fibres per $\text{mm}^2$), four current source propagation velocities (3.0, 3.5, 4.0, 4.5 m/s), four IZ positions in ratio (0.4, 0.46, 0.53, 0.6), and four fibre lengths in ratio (0.85, 0.95, 1.05, 1.15). The other two specified system conditions are the depth and the medial-lateral position of the MU centre, which was defined as the geometric centre of all the muscle fibres it controlled. The six specified system conditions were linearly normalised between 0.5 and 1 and concatenated to provide a specified condition $\boldsymbol{c_s}$ for each $\boldsymbol{x}$.

Each MUAP template consisted of a $10 \times 32$ electrode grid with a time support of 96 samples at a sampling frequency of 2048Hz (48ms). There were a total of 384,000 samples in the dataset, which were then divided into training and validation datasets. All training was completed using the training dataset, with the validation dataset only used to test the model after all parameter and hyperparameter optimisation had been completed.

\subsection{Network Architecture}
The architectures of the encoder in the generator and the discriminator are described in Table \ref{tbl:nn}. The encoder projects a batch of input data $\boldsymbol{x} \in \mathcal{R}^{1 \times T \times W \times H}$ to the encoded variables
$\boldsymbol{c_u} \in \mathcal{R}^{C \times T' \times W' \times H'}$
through five convolutional layers. Each convolutional layer consists of a 3D convolution with kernel size 3 and a $1 \times 1$ convolution with a skip connection. A parametric rectified linear unit (PReLU) activation function is placed after each convolution. The output of the convolutional layers is then passed to two linear layers, which estimate the statistics of the approximate posterior, the mean and variance, respectively. During training, the approximate posterior of the encoded MUAP template was estimated from the statistics by the reparameterisation trick. During inference, the expectation was used for transforming existing data while for \textit{ab initio} generation, the approximate posterior was sampled from the prior distribution. Before passed to the decoder, the posterior was concatenated with the representation of the specified system conditions, which was projected by a linear layer.

\begin{table}[h]
  \captionsetup{justification=centering}
  \caption{The network structures of encoder and discriminator in \BioMime/.}
  \begin{center}
      \resizebox{\linewidth}{!}
      {
          \begin{tabular}{cccccc}
              \toprule
              \multicolumn{3}{c}{Generator-Encoder} & \multicolumn{3}{c}{Discriminator} \\
              \midrule
              Layer & Parameters & Output & Layer & Parameters & Output\\
              \midrule
              Conv3d & 3, 2, 1 & [16, 48, 5, 16] & Conv3d & 3, 1, 1 & [16, 48, 5, 16] \\
              Conv3d & 3, 2, 1 & [32, 24, 3, 8] & Conv3d & 3, 2, 1 & [32, 24, 3, 8] \\
              Conv3d & 3, 2, 1 & [64, 12, 2, 4] & Conv3d & 3, 2, 1 & [64, 12, 2, 4] \\
              Conv3d & 3, (2, 1, 1), 1 & [128, 6, 2, 4] & Conv3d & 3, (2, 1, 1), 1 & [128, 6, 2, 4] \\
              Conv3d & 3, 1, 1 & [256, 6, 2, 4] & Conv3d & 3, 1, 1 & [256, 6, 2, 4] \\
              \midrule
              Flatten & ~ & [256 $\times$ 6 $\times$ 2 $\times$ 4] & AvgPool3d & (6, 2, 4) & [256, 1, 1, 1]\\
              Linear ($\mu$) & 12288 $\rightarrow$ 16 & [16] & Conv3d & 1, 1, 0 & [1, 1, 1, 1]\\
              Linear ($\sigma$) & 12288 $\rightarrow$ 16 & [16] & Squeeze & ~ & [1]\\
              \bottomrule
          \end{tabular}
      }
  \end{center}
  \footnotesize{$^*$ Each Conv3d in the encoder is followed by a PReLU layer. Each Conv3d in the discriminator is followed by a LeakyReLU with negative slope equal to 0.03.}\\
  \label{tbl:nn}
\end{table}

\begin{figure}[h]%
  \centering
  \includegraphics[width=0.45\textwidth]{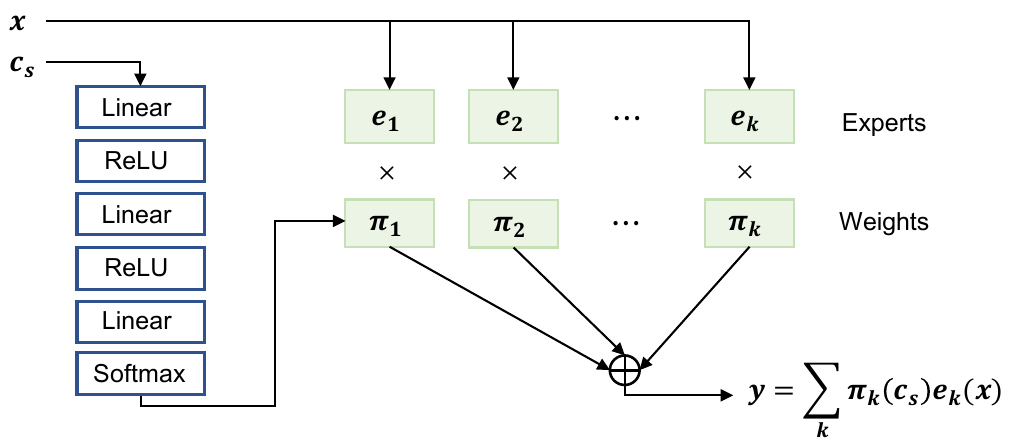}
  \caption[Time-scaling module in decoder]{Time-scaling module in decoder. The proposed time scaling module is a temporal pooling/unpooling bank with learned weights. The weights are conditioned on the desired specified system conditions $\boldsymbol{c_s}$ by a feedforward layer. This module is specialised for dilating or compressing signals in time domain.
  }\label{fig:expert}
\end{figure}

The decoder aims to generate new data $\boldsymbol{\tilde{x}}$ from the concatenated representations of the specified and unspecified conditions, $\boldsymbol{c_s}$ and $\boldsymbol{c_u}$. It consists of four convolutional layers with PReLU activations, followed by two rescaling blocks (sequentially a time-scaling module, a 3D convolution, and a $1 \times 1$ convolution). The time-scaling module proposed in this study was used to flexibly transform the latent features back to MUAP templates. The time-scaling module is a mean pooling\textbackslash unpooling bank with learned weights, specialised for rescaling time sequences in a large scale, which is essential in modelling volume conductor effects while difficult to achieve by piling a limited number of convolutional layers. Inspired by the dynamic convolution \cite{chen2020dynamic} and the multi-scale spatial pooling \cite{chen2020convolutional}, the time-scaling module uses a series of experts $e_k$ to rescale the input signals with different scaling factors $s_k$. A factor $s_k > 1.0$ means dilating the signal while $s_k < 1.0$ means compressing the signal. Since the scale of the expected MUAP relies on the specified system conditions, the weights of the experts are projected from the specified system conditions by a feedforward neural network (three linear layers with two activation layers) and are normalised by softmax (Fig. \ref{fig:expert}).
The compressed signals are padded with zeros at the end while the dilated ones are truncated to have the same size. The outputs of the experts are aggregated by:

\begin{equation}
  \begin{aligned}
    &\boldsymbol{y} = \sum_k \pi_k(\boldsymbol{c}) e_k(\boldsymbol{x}) \\
    &s.t. 0 \leq \pi_k(\boldsymbol{c}) \leq 1, \sum_{k=1}^{K} \pi_k(\boldsymbol{c}) = 1
  \end{aligned}
\end{equation}
where $\pi_k(\boldsymbol{c})$ is the weight of the kth expert conditioned on the desired specified system conditions $\boldsymbol{c_s}$.

As $\pi_k$ is projected from the conditions of each individual sample and jointly optimised with the network, the model automatically identifies significant experts and dynamically upscales the data in a non-linear way. In this study, we used eight experts with scaling factors linearly spaced between 0.25 and 1.0, which show accurate generation within an acceptable number of training iterations.

The discriminator in \BioMime/ has the same convolutional layers with the encoder in generator to extract the features of the inputs. The system conditions are concatenated with the output of the first shortcut \cite{perarnau2016invertible}. Finally, the features are averaged among the temporal and spatial dimensions and projected to a scalar by a convolutional layer.

\subsection{Training Pipeline}

The full adversarial training algorithm is described in algorithm \ref{alg:training}. The generator and discriminator were trained using the RMSprop optimiser with a learning rate of ${1 \times 10^{-5}}$. The networks were trained for 45 epochs (405,000 iterations), taking 120 hours on an NVIDIA RTX 2080Ti. The ratio between the updates of parameters of discriminator and generator was $1 : 1$. The hyperparameters were selected by grid search, with $\lambda_1$ set to 10, $\lambda_2$ an annealing weight from 0 to 0.05 over 30,000 iterations\cite{bowman2015generating}, $\lambda_3$ set to 0.5, $\lambda_4$ set to 1.0, $\lambda_5$ set to 0.5 and $\lambda_6$ set to 0.5.

\begin{algorithm}
  \caption{Training procedure of \BioMime/}\label{alg:training}
  \begin{algorithmic}[1]
      \Require MUAP samples labelled with their specified system paramters \{$\boldsymbol{x_0}, \boldsymbol{c_0}$\}, the desired conditions $\boldsymbol{c_{s}}$, randomly sampled system conditions $\boldsymbol{c_{r}}$, number of training epochs $S$, hyperparameters $\lambda_1 = 10.0$ for $\mathcal{L}_{GAN}$, \textit{anneal\_func} for linearly increasing the weight $\lambda_2$ of $\mathcal{L}_{KL}$, $\lambda_3 = 0.5$ for $\mathcal{L}_{cyclic}$
  \State $\boldsymbol{\theta_{G}}, \boldsymbol{\theta_{D}}$ $\leftarrow$ initialize network parameters
  \For{$n = 1$ to $S$}
      \State $\rho_r \leftarrow D(\boldsymbol{x_0}, \boldsymbol{c_0})$ \quad\quad \#{ real sample + real condition}
      \State $\rho_{f1} \leftarrow D(G(\boldsymbol{x_0}, \boldsymbol{c_{s}}), \boldsymbol{c_{s}})$ \ \ \#{ fake sample + real condition}
      \State $\rho_{f2} \leftarrow D(\boldsymbol{x_0}, \boldsymbol{c_{r}})$  \ {\# real sample + random condition}
      \State $\mathcal{L}_D \leftarrow \lambda_4(1 - \rho_r)^2 + \lambda_5\rho_{f1}^2 + \lambda_6\rho_{f2}^2$
      \State $\boldsymbol{\theta_D} \leftarrow \boldsymbol{\theta_D }- \alpha \boldsymbol{\nabla}_{\theta_D}\mathcal{L}_D$ \quad \ \ \ {\# update discriminator}
      \State $\boldsymbol{\tilde{x}}, \boldsymbol{\mu}, log(\sigma^2) \leftarrow G(\boldsymbol{x_0}, \boldsymbol{c_s})$  \ {\# generate by morphing}
      \State ${\boldsymbol{\hat{x}}} \leftarrow G(\boldsymbol{\tilde{x}}, \boldsymbol{c_0})$ \quad\quad\quad \quad \quad \ {\# reverse generation}
      \State $\rho \leftarrow D(\boldsymbol{\tilde{x}}, \boldsymbol{c_s})$
      \State $\mathcal{L}_{GAN} \leftarrow (1 - \rho)^2$
      \State $\mathcal{L}_{cyclic} \leftarrow \|\boldsymbol{x_s} - \boldsymbol{\hat{x}}\|_2^2$
      \State $\mathcal{L}_{KL} \leftarrow -(1+log(\sigma^2) - \mu^2 - \sigma^2) / 2$
      \State $\lambda_2 \leftarrow anneal\_func(n)$
      \State $\mathcal{L}_G \leftarrow \lambda_1 \mathcal{L}_{GAN} + \lambda_2 \mathcal{L}_{KL} + \lambda_3 \mathcal{L}_{cyclic}$
      \State $\boldsymbol{\theta_G} \leftarrow \boldsymbol{\theta_G} - \alpha \boldsymbol{\nabla}_{\theta_G}\mathcal{L}_G$ \quad\quad\quad \quad \ \ \ {\# update generator}
  \EndFor
  \end{algorithmic}
\end{algorithm}

\subsection{Analysis Methods}
To validate the model, we evaluated the ability of BioMime to transform the existing MUAP templates to new sets of simulation parameters. This was performed by optimising the model on the training dataset and evaluating its performance on the held-out validation set. We randomly selected two data samples with the same unspecified generative factors and transformed the first sample to match the second sample's specified system conditions. Accuracy was evaluated by the normalised root mean square (nRMSE) between the predicted MUAP and the ground truth MUAP:

\begin{equation}
  \text{nRMSE}=\frac{\sqrt{\sum_{h, w, t}\left(x_{h, w, t}-\tilde{x}_{h, w, t}\right)^{2} / H / W / T}}{x_{max} - x_{min}}
\end{equation}
where $x$ and $\tilde{x}$ indicate the ground truth MUAP from the numerical model and the synthesised MUAP by BioMime. The variables $h$, $w$, and $t$ are the summation over the rows and columns of the electrode and the time samples, with the total number $H$, $W$, and $T$, respectively. $x_{max}$ and $x_{min}$ are the maximum and minimum values of the sample. The nRMSE is often expressed as a percentage, with a lower value indicating a less residual variance for the model and a higher generation accuracy. We empirically found that an nRMSE less than 2.0\% indicates a good generation with few differences between the generated data and the ground truth data.

The ability of BioMime to interpolate between system conditions was visualised by using t-distributed stochastic neighbor embedding (t-SNE). The MUAP templates from both the FEM dataset and BioMime were embedded. This was implemented using scikit-learn's t-SNE module with default settings (dimension of the embedded space 2, perplexity 30, learning rate 200, maximum number of iterations 1000, initialisation of embedding using principal component analysis) \cite{scikit_learn}. The samples were visualised in the coordinates given by the dimension reduction.

\begin{figure*}[htb]
  \centering
  \includegraphics[width=1.85\columnwidth]{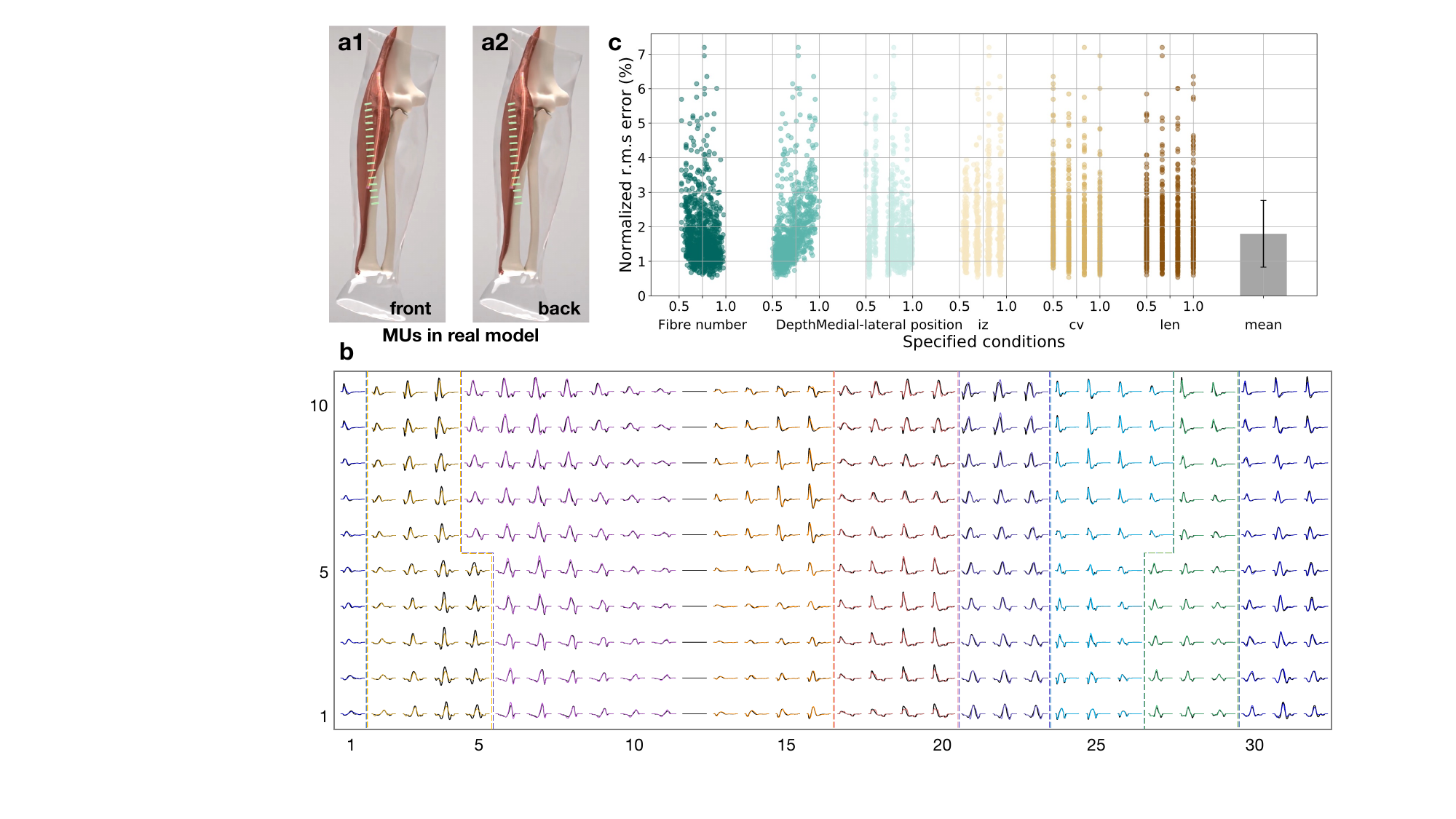}
  \caption[\BioMime/ accurately learns to mimic a biophysical simulation]{\BioMime/ accurately learns to mimic a biophysical simulation. a, Muscle fibre locations of the eight representative MUs highlighted in different colors from front and back. Each MU is from a separate muscle in the realistic forearm volume conductor. The black dots represent the surface electrode array with $10 \times 32$ channels and the inter-electrode distance is 8 mm. b, MUAP signals synthesised by conditionally transforming the MUAPs of the eight MUs in a, compared with the ground truth in black lines. For each MU, only channels with large amplitudes are illustrated in sub-grids. These channels represent the electrodes that are physically closest to the MU. Sub-grids from different MUs are separated by dotted lines. The numbers beside and below the grid denote the indices of electrodes in row and in column, respectively. The shape of each transformed MUAP from \BioMime/ closely matches its simulated counterpart.
  c, Normalised root mean square error in percentage between the outputs from \BioMime/ and those from the numerical simulator across the six specified system conditions and overall evaluated on the validation set. The normalised range of each condition's values is represented on the x-axis. The effect of all of the specified system conditions on the simulated outputs was closely mimicked by \BioMime/.
  }
  \label{fig:accurate}

\end{figure*}

We also investigated whether concatenating the encoded MUAP templates $\boldsymbol{c_u}$ with the specified system conditions $\boldsymbol{c_s}$ causes the encoder to ignore the variance in the templates explained by $\boldsymbol{c_s}$, i.e. whether it leads to a semi-supervised latent disentanglement\cite{gyawali2019semi, wang2021self}. We expect this would improve the predictability of the model to the changes in the specified system conditions by reducing the burden on the decoder. This effect was investigated quantitatively using an informativeness metric \cite{eastwood2018framework}, which measures how much information of the specified system conditions is contained within the latent features. Specifically, the informativeness of the latent vector $\boldsymbol{c_u}$ about the $i$th generative factor $\boldsymbol{c_{s_i}}$ can be quantified by the prediction accuracy $P(\boldsymbol{c_{s_i}}, f(\boldsymbol{c_{u_i}}))$, where $P$ is an accuracy metric and $f$ is a regressor \cite{eastwood2018framework}. The informativeness depends on (1) the regressor's capability to extract information from the latent representations and (2) the way the specified system conditions are embedded in the latent features, e.g., disentanglement of the specified system conditions in the representations will make the regression easier. We used a multilayer perceptron with non-linear activation functions to predict $\boldsymbol{c_{s_i}}$ from $\boldsymbol{c_{u_i}}$. The prediction accuracy was measured as the percentage of the correctly predicted samples.

\subsection{Simulating Naturalistic Movements}
As an example of potential applications, BioMime was used to simulate surface EMG during dynamic contractions of a naturalistic movement.
We selected the movement of opening and closing the hand whilst flexing and extending the wrist (Fig. \ref{fig:msk}a). The movement was modelled in OpenSim\cite{seth2018opensim, del2017associations}, using the ARMs Wrist and Hand Model \cite{mcfarland2022musculoskeletal}. The muscle fibre lengths, one of the system conditions, were then estimated by using the Muscle Analysis Tool in OpenSim.

BioMime was used in its generative mode to simulate 1,500 MUAP templates across eight muscles, using the initial system conditions in the time series. It was then used in its morphing mode to modify each of the MUAP templates as the system conditions varied along the time series. Motor unit spike trains were then generated using a contemporary model of motor recruitment with a trapezoidal neural input \cite{fuglevand1993models}. The MUAP templates were convolved with their respective motor unit spike trains to obtain the individual source contributions, which were summed with Gaussian-distributed noise to produce the final surface EMG signals.

\section{Results}
\subsection{Model Accuracy}
BioMime was able to consistently and accurately transform MUAP templates based on the specified system conditions and generate samples that were extremely similar in shape to the equivalent of the numerical outputs (Fig \ref{fig:accurate}). Only slight deviations at the amplitudes and end-fibre effects appeared in a few electrode channels. The mean nRMSE between all samples and their FEM equivalents was 1.8\% (std 1.0\%) when evaluated on the held-out validation set. There was relatively equal performance across all six of the specified system conditions.

\subsection{Predictive Interpolation}
An important feature of BioMime is its ability to simulate new data by interpolating between the system conditions of the FEM models used for training, thereby reducing the need for FEM runs.
We selected four specified system conditions $\boldsymbol{c_{s_i}}, i = 0, 1, 2, 3,$ for the samples in the validation dataset, such that their associated templates were the result of a linear sweep through the specified conditions. The first template $\boldsymbol{x_0}$ with $\boldsymbol{c_{s_0}}$ was encoded to give the $\boldsymbol{c_u}$ for that MUAP. When we simulate a sequence of MUAP templates by concatenating $\boldsymbol{c_u}$ with the interpolated specified system conditions from $\boldsymbol{c_{s_0}}$ to $\boldsymbol{c_{s_3}}$, the output MUAPs gradually hit the paired ground truth templates in the validation dataset, with which the interpolation accuracy was evaluated.

\begin{figure}[h]
  \centering
  \includegraphics[width=0.9\columnwidth]{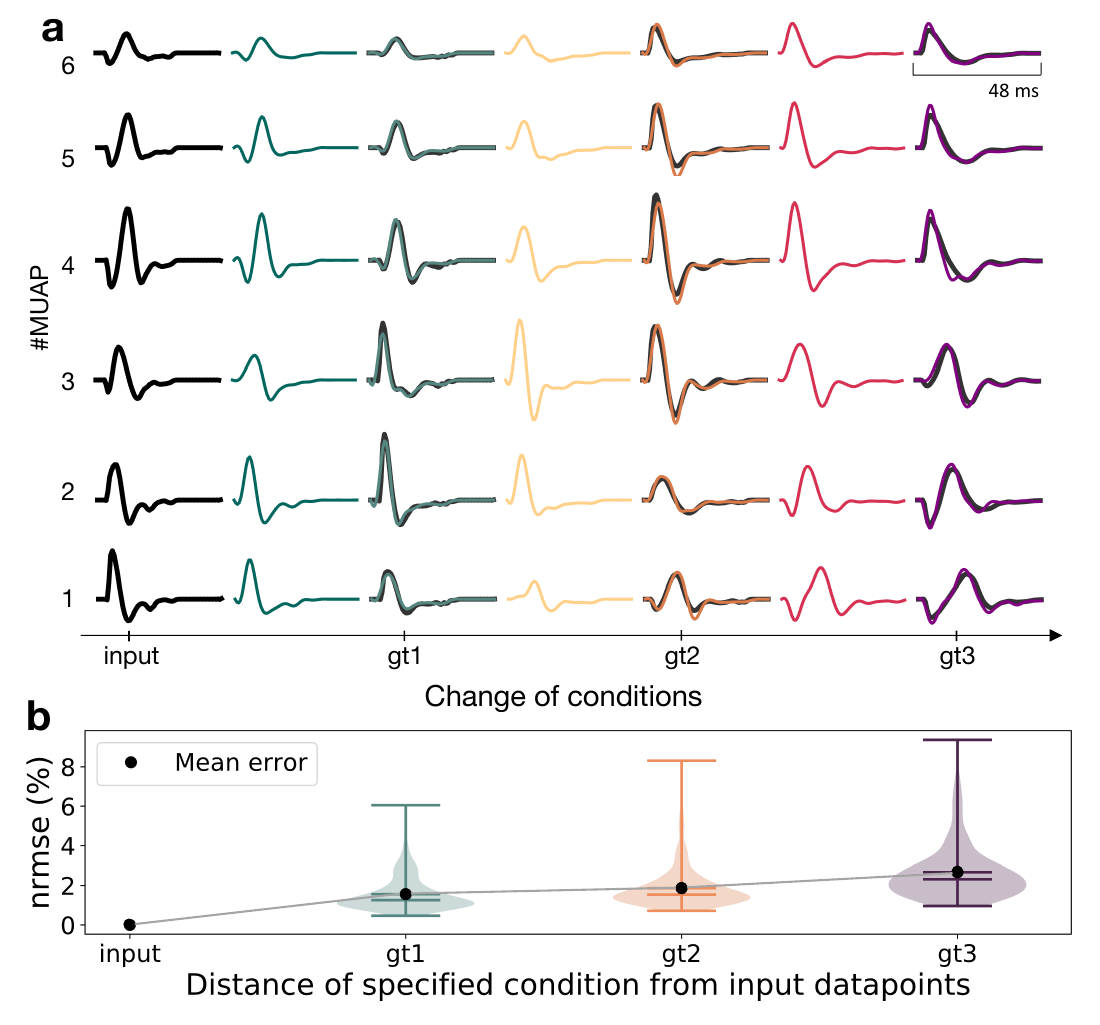}
  \caption[BioMime can predictively interpolate between simulation conditions and beyond]{BioMime can predictively interpolate between simulation conditions and beyond.
  a, A set of BioMime-transformed MUAPs sampled from a continuous sweep of specified system conditions. The original MUAP templates in the validation dataset were encoded and then continuously transformed away from their origins (x-axis tick label of `input'). As the specified system conditions were continuously traversed, they occasionally intersected with the generative factors from the numerical simulator (displayed as superimposed black lines, x-axis tick labels of `gt1', `gt2', and `gt3'), which can be used as a ground truth.
  b, Mean normalised root mean squared error of the BioMime-transformed MUAPs compared to their ground truth counterparts from the validation set during the traversal in a.
  The predictive error of BioMime increased as the specified system conditions moved further away from their origin. BioMime is able to compensate for large moves away from the original conditions before the error starts to rise.
  }
  \label{fig:sweep}
\end{figure}

As $\boldsymbol{c_{s_i}}$ moved further away from the original $\boldsymbol{c_{s_0}}$, the MUAP shape deviates away from the ground truth (Fig. \ref{fig:sweep}). However, the discrepancy was not large until the sweep intersected with the third ground truth, which was the sample with the specified system conditions most different from the original point. On the sweep intersections with the first, second, and third ground truths, the mean nRMSEs were respectively 1.8\% (std 0.6\%), 1.9\% (std 0.7\%), and 3.6\% (std 1.1\%). This means that even for situations where very high interpolation accuracies are desired during a dynamic traversal, a relatively small dataset from the computational expensive numerical simulator is sufficient.

\begin{figure}[h]
  \centering
  \includegraphics[width=\columnwidth]{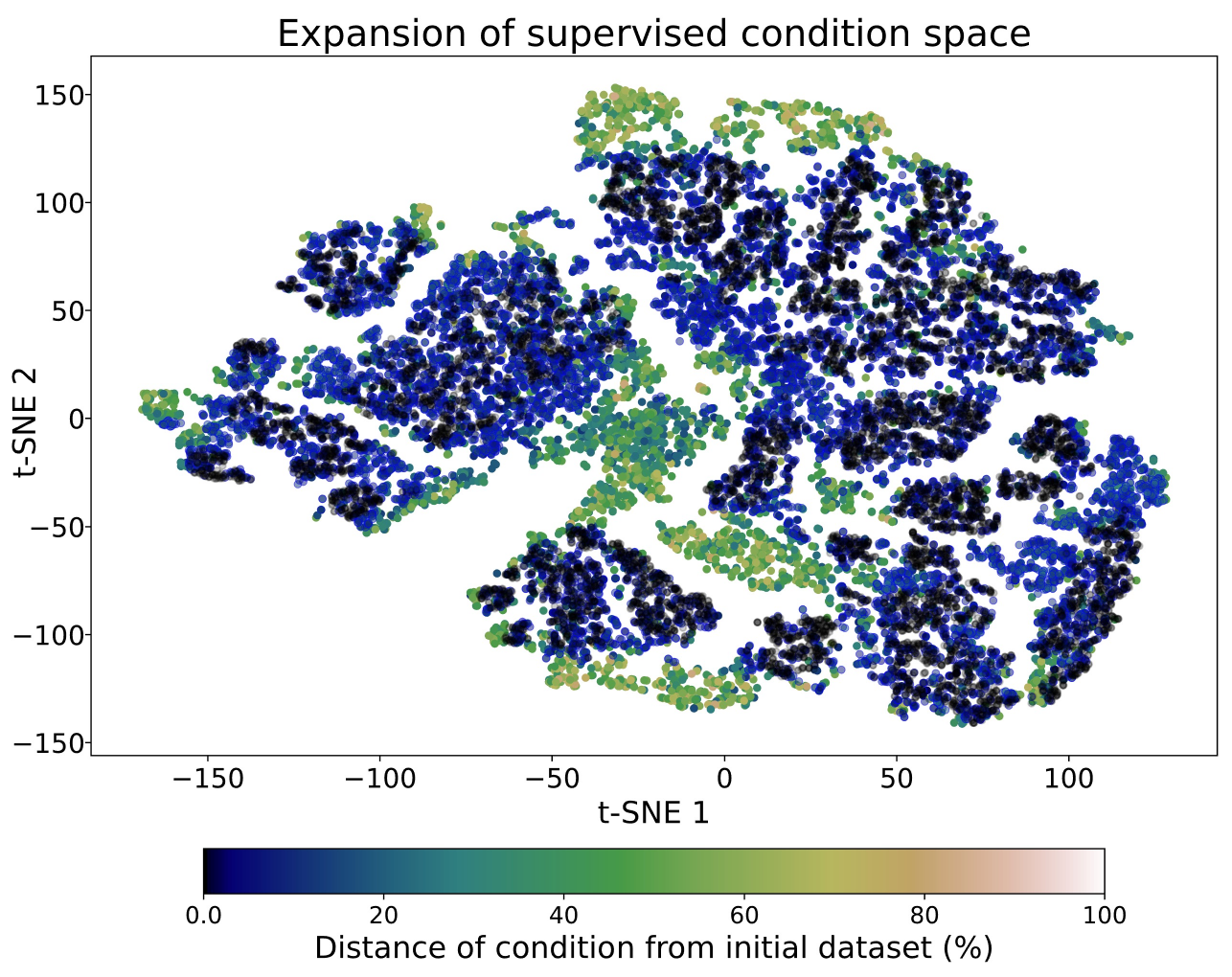}
  \caption[Distribution of MUAPs generated by the numerical model and by BioMime]{A t-SNE projection of the MUAP templates from the FEM model (black) and from BioMime. The BioMime templates are gradient-coloured by their distance from their nearest numerical counterparts. The distribution of the interpolated MUAPs from BioMime closely matched those from the FEM model.
  }
  \label{fig:tsne}
\end{figure}

When embedded into two dimensions by t-SNE, the samples in the original dataset were grouped into small clusters (Fig. \ref{fig:tsne}), since the MUAPs from the same muscle are highly similar. The data generated by interpolating the specified system conditions closely matched the original data in the low dimensional space, while the others with extrapolated specified system conditions only started to deviate until the specified system conditions were different from the original conditions by a relative difference of 60\%. This indicates that BioMime is able to expand the data space whilst retaining generation accuracy.

\subsection{Latent Disentanglement}
We found that the latent features held very little information about the specified system conditions, with a median informativeness score slightly above chance at $35.9\%$. The low informativeness metric indicates that the generative effects of the specified system conditions were disentangled from those that were unspecified, i.e., not explicitly included during training. This means that when new MUAP templates are generated by morphing from an existing template, the specified system conditions of the input MUAP will not overwrite the new system conditions that are explicitly provided.

\subsection{Ablation Analysis of the Time-scaling Module}
The decoder in BioMime aims to generate new data that has the same high dimensions as the input. A na\"ive implementation is to use 3D fractionally-strided convolutions, which are commonly used to restore images. However, this implementation yielded low generation accuracy. One possible reason is that simulating physiological signals requires the model to flexibly transform sequences with long temporal dynamics, with which traditional convolution filters have been noted to struggle\cite{bai2018empirical}. The ablated model had inferior performance to the model that used the time-scaling module in the encoder, with a mean nRMSE of 4.2\% (std 1.2\%) for the ablated model and a mean nRMSE of 2.1\% (std 1.1\%) for the model using the time-scaling module.

We found that ranging the temporal scaling from 0.25 to 2.0 is sufficient for flexibly transforming the MUAP signals of human forearm muscles. Increasing the number of experts in the time-scaling modules introduces more scaling factors but also increases the training time. A time-scaling module with eight experts generated signals with high accuracy and in an acceptable number of training iterations and was selected for the final architecture

\subsection{Rapid Simulation of sEMG During Naturalistic Movements}

\begin{figure*}[htb]
  \centering
  \includegraphics[width=0.95\textwidth]{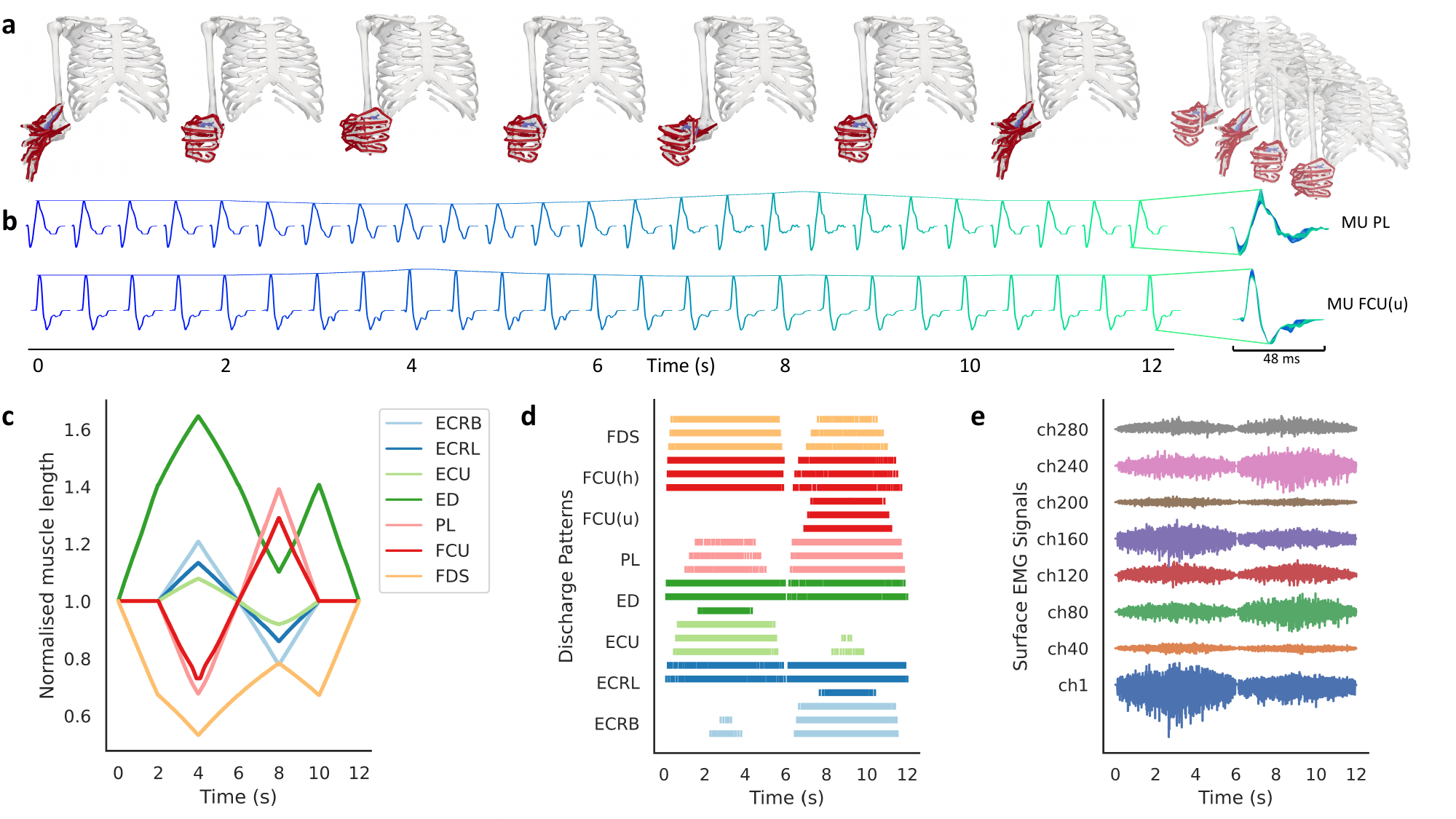}
  \caption[Combining BioMime with musculoskeletal and motor neuron modelling]{BioMime can be used to mimic the dynamic changes of a biophysical system that matches the mechanical movements of a musculoskeletal model.
  a, Sequence of the movements. PE, palm extension; HG, hand grasp; WF, wrist flexion; WE, wrist extension.
  b, Two representative MUAP templates generated by BioMime, which are continuously morphed using the muscle fibre lengths given by the musculoskeletal model. The upper MUAP is from Palmaris longus, PL and the lower MUAP is from the ulnar head of Flexor carpi ulnaris, FCU(u). The lines that connect the peak of MUAPs show the changes in MUAP amplitude. The MUAPs do not show extreme variations. This is expected as only one specified condition, the muscle fibre length, is changed during the transformation.
  c, Normalised muscle fibre lengths from the musculoskeletal model. ECRB, Extensor carpi radialis brevis; ECRL, Extensor carpi radialis longus; ECU, Extensor carpi ulnaris; ED, Extensor digitorum; PL, Palmaris longus; FCU, Flexor carpi ulnaris (ulnar head and humeral head); FDS, Flexor digitorum superficialis.
  d, Representative discharge patterns of three MUs in each muscle. The motor unit spike trains were produced by the motoneuron pool model in \cite{fuglevand1993models} by simulating a trapezoidal activation of the flexors and the extensors. The activations of the flexor group (FDS, FCU(h), FCU(u), PL) and the extensor group (ED, ECU, ECRL, ECRB) are displayed.
  e, Simulated surface EMG signals in eight channels. This complex dynamic simulation is only possible due to the low computational cost of the hundreds of MUAP template transformations conducted by BioMime.}
  \label{fig:msk}
\end{figure*}

Lastly, we provide the output of a case study in which BioMime was used to simulate the EMG signal detected by surface electrodes during a naturalistic movement.
We simulated a hand and wrist movement with the ARMs Wrist and Hand Model \cite{mcfarland2022musculoskeletal} and estimated the muscle fibre lengths during the movement using OpenSim\cite{seth2018opensim, del2017associations}. The spike trains were generated by a well-developed motoneuron pool model \cite{fuglevand1993models}, with a trapezoidal neural input. The library of the transformed MUAPs was then convolved with the motor neuron spike trains to give surface EMG (Fig. \ref{fig:msk}).
Such a simulation would be prohibitively time-consuming using numerical methods due to the complicated pre-processing steps and the computational cost of simulating hundreds of MUAPs at discretised stages of the movement. By contrast, BioMime's ability to synthesise MUAP templates from specified system conditions at any arbitrary level of time resolution made such a simulation possible. The MUAP template waveforms changed smoothly and continuously during the movement. The duration of the waveform and IZ changed the most during the movement, which is expected because they are directly affected by the muscle fibre length.

\section{Discussion}
Humans are by their nature dynamic systems. Computationally-feasible methods are needed to better reflect this fact in biophysical simulations such as in EMG generation. By transferring the knowledge contained within a set of FEM simulations to a conditional generative model, we demonstrate that the cost associated with simulating an evolving biophysical system can be largely mitigated without a significant drop in predictive fidelity. With the correct architecture design, such models can double both as a rapid way to transform existing data to reflect new system conditions and as \textit{de novo} generators for new outputs.

BioMime, the proposed encoder-decoder neural network trained adversarially using a conditional discriminator, was found to be an excellent platform for learning to mimic a large set of FEM simulation outputs.
BioMime is able to accurately interpolate between the system conditions of a set of numerical simulators, thereby greatly reducing the required number of mesh changes and forward model recalculations. Consequently, we were able to demonstrate the first practical application for simulating the myoelectric output of a moving forearm during a realistic hand and wrist movement. This represents a step forward in terms of surface EMG simulation which we hope will have an immediate impact on downstream research in surface EMG decomposition \cite{glaser2018motor}, inverse modelling, and human-machine interface design.

The model is easily controlled through the concatenation of a simplified set of specified system conditions without sacrificing the variance explained by the other components of the complex numerical model. Such a design also encourages semi-supervised latent disentanglement, which reduces the burden of the decoder. The architecture of the decoder, particularly the novel time-scaling module, was found to be highly suitable for recreating MUAP templates with a high level of generalisability. We expect such modules to offer new ways in processing physiological signals using deep learning methods.

It is important to emphasise that the methods outlined in this paper are designed to augment and approximate, rather than replace the high-quality numerical modelling. In studies of peripheral motor systems, the dynamic changes in system conditions in response to some perturbations remain poorly characterised, which in turn restricts the utility of FEM models and the proposed generative model. This is particularly true when modelling dynamic changes in a volume conductor for generating surface EMG signals. While we attempted to carefully estimate the likely parameter changes during a forearm movement by using an advanced musculoskeletal model, only muscle fibre lengths can be tracked, which limits the precise prediction of MUAP template changes. There is also a dearth of such information in the literature. We hope that the promise offered by conditional generative models in capturing the dynamic systems will further stimulate research in this direction, which will in turn improve the generation accuracy of models such as BioMime.

In conclusion, we present a conditional generative model that extends existing advanced numerical models of surface EMG generation, such that the computational cost of simulating naturalistic movements is not prohibitively expensive. We demonstrate the ability of the conditional generative model to learn the combined output of many FEM simulations of volume conduction and to accurately interpolate between them. Such accuracy and efficiency dramatically reduce the number of FEM runs needed in order to have a smooth change in the biophysical outputs when the system anatomy evolves. We anticipate that the pipeline of a back-end teacher numerical model and a front-end student generative model will become increasingly common in future simulation design, and we look forward to the corresponding expansion in dynamic simulations that these techniques will enable.

\section{Code availability}
All code was implemented in Python using the deep learning framework PyTorch. Code which implements the models used in this paper is available at \href{https://github.com/shihan-ma/BioMime}{https://github.com/shihan-ma/BioMime} and is provided under the GNU General Public License v3.0.

\section{Acknowledgements}
We would like to thank Pranav Mamidanna for the discussions and assistance in preparing the musculoskeletal model.

\bibliographystyle{ieeetr}
\bibliography{bibliography}

\end{document}